\pdfoutput=1

\documentclass[11pt]{article}

\newcommand{\draftonly}[1]{#1}
\renewcommand{\draftonly}[1]{}

\usepackage[final]{acl}

\usepackage{times}
\usepackage{latexsym}

\usepackage[T1]{fontenc}

\usepackage[utf8]{inputenc}

\usepackage{microtype}

\usepackage{inconsolata}

\usepackage{graphicx}

\RequirePackage{algorithm}
\RequirePackage{algorithmic}

\usepackage{microtype}

\usepackage{amssymb,bm,mathtools,color}

\usepackage[normalem]{ulem}
\usepackage{enumitem}

\usepackage{booktabs}
\usepackage{multirow}
\usepackage{tabularx}
\usepackage{multicol}
\usepackage{multirow}
\usepackage{colortbl}
\usepackage{hhline}
\usepackage{stfloats}

\newcolumntype{Y}{>{\centering\arraybackslash}X}

\usepackage{xcolor}
\definecolor{red}{rgb}{0.74,0.08,0.10}
\definecolor{green}{rgb}{0.26,0.49,0.18}
\definecolor{blue}{rgb}{0.22,0.53,0.75}

\usepackage{colortbl}
\definecolor{Gray}{gray}{0.9}
\definecolor{LightCyan}{rgb}{0.75,1,1}

\makeatletter
\newcommand\notsotiny{\@setfontsize\notsotiny\@viiipt\@ixpt}
\makeatother
\usepackage{booktabs}
\usepackage{multirow}
\usepackage{graphicx}

\usepackage{pifont}
\usepackage{amsmath}
\usepackage{amssymb}
\usepackage{ifthen}
\usepackage{dsfont}
\usepackage{bold-extra}
\usepackage{courier}

\usepackage{listings}
\usepackage{xcolor}
\usepackage{ltablex}
\usepackage{longtable}
\usepackage{multicol}

\usepackage{tikz}
\usetikzlibrary{bayesnet}
\usepackage[linguistics]{forest}

\usepackage{caption}
\usepackage{cancel}

\usepackage{xspace}
\usepackage{comment}
\usepackage{color,soul}

\usepackage{float}

\usepackage[noabbrev,capitalize]{cleveref} 

\crefname{page}{page}{pages}
\crefname{footnote}{footnote}{footnotes}   
\crefname{equation}{equation}{equations}   

\creflabelformat{equation}{#2\textup{#1}#3}

\crefname{line}{line}{lines}               
\crefrangeformat{line}{lines #3#1#4--#5#2#6}
\crefname{lstlsting}{Listing}{Listings}   
\crefname{section}{\S}{\S\S}
\Crefname{section}{\S}{\S\S}    
\crefformat{section}{#2\S#1#3}  
\Crefformat{section}{#2\S#1#3}
\crefrangeformat{section}{\S\S#3#1#4--#5#2#6}
\Crefrangeformat{section}{\S\S#3#1#4--#5#2#6}
\crefmultiformat{section}{\S#2#1#3}{ and~\S#2#1#3}{, \S#2#1#3}{ and~\S#2#1#3}
\Crefmultiformat{section}{\S#2#1#3}{ and~\S#2#1#3}{, \S#2#1#3}{ and~\S#2#1#3}
\crefrangemultiformat{section}{\S\S#3#1#4--#5#2#6}{ and~\S\S#3#1#4--#5#2#6}{, \S\S#3#1#4--#5#2#6}{ and~\S\S#3#1#4--#5#2#6}
\Crefrangemultiformat{section}{\S\S#3#1#4--#5#2#6}{ and~\S\S#3#1#4--#5#2#6}{, \S\S#3#1#4--#5#2#6}{ and~\S\S#3#1#4--#5#2#6}

\DeclareMathOperator*{\argmax}{arg\,max}
\newcommand{\prob}[2][]{p\ifthenelse{\not\equal{}{#1}}{_{#1}}{}(#2)} 
\newcommand{\expect}[2][]{\text{\bf E}\ifthenelse{\not\equal{}{#1}}{_{#1}}{}\!\left[#2\right]}
\newcommand{\var}[2][]{\text{\bf Var}\ifthenelse{\not\equal{}{#1}}{_{#1}}{}\!\left[#2\right]}

\newcommand{\ie}{\emph{i.e.}, }
\newcommand{\eg}{\emph{e.g.}, }

\newcommand{\makename}[3][s]{%
  \expandafter\newcommand\csname #2\endcsname{#3\xspace}%
  \expandafter\newcommand\csname #2s\endcsname{#3#1\xspace}%
}

\makename{aucmse}{\texttt{AUC-MSE}}
\makename{scauc}{\texttt{SCAUC}}

\usepackage[]{todonotes}

\usepackage[]{todonotes}  
\makeatletter
\newcommand*\iftodonotes{\if@todonotes@disabled\expandafter\@secondoftwo\else\expandafter\@firstoftwo\fi}  
\makeatother

\title{SimBA: Simplifying Benchmark Analysis\\Using Performance Matrices Alone}
\author{\
Nishant Subramani$^{\text{\ding{171}}}\thanks{Equal Contribution}$%
\ \ \ \ \ 
\textbf{Alfredo Gomez$^{\text{\ding{171}}*}$}%
\ \ \ \ \ 
\textbf{Mona Diab$^\text{\ding{171}}$}%
\\
\\$^\text{\ding{171}}$Carnegie Mellon University - Language Technologies Institute \\
$\texttt{\{nishant2,alfredo3,mdiab\}@cs.cmu.edu}$ \\
}

\begin{document}
\maketitle
\frenchspacing

\begin{abstract}
Modern language models are evaluated on large benchmarks, which are difficult to make sense of, especially for model selection.
Looking at the raw evaluation numbers themselves using a model-centric lens, we propose \textbf{SimBA}, a three phase framework to \textbf{Sim}plify \textbf{B}enchmark \textbf{A}nalysis.
The three phases of \textbf{SimBA} are: \texttt{stalk}, where we conduct dataset \& model comparisons, \texttt{prowl}, where we discover a representative subset, and \texttt{pounce}, where we use the representative subset to predict performance on a held-out set of models. 
Applying \textbf{SimBA} to three popular LM benchmarks: HELM, MMLU, and BigBenchLite reveals that across all three benchmarks, datasets and models relate strongly to one another (\texttt{stalk}).
We develop an representative set discovery algorithm which \emph{covers} a benchmark using raw evaluation scores alone.
Using our algorithm, we find that with 6.25\% (1/16), 1.7\% (1/58), and 28.4\% (21/74) of the datasets for HELM, MMLU, and BigBenchLite respectively, we achieve coverage levels of at least 95\% (\texttt{prowl}).
Additionally, using just these representative subsets, we can both preserve model ranks and predict performance on a held-out set of models with \emph{near zero} mean-squared error (\texttt{pounce}). 
Taken together, \textbf{SimBA} can help model developers improve efficiency during model training and dataset creators validate whether their newly created dataset differs from existing datasets in a benchmark. Our code is open source, available at \url{https://github.com/nishantsubramani/simba}.
\end{abstract}
\section{Introduction}
The rapid expansion of language model (LM) benchmarks has resulted in an overabundance of evaluation datasets. 
However, the relationships among these datasets remain poorly understood.
Current evaluation methods primarily focus on overall model win rates or simple aggregate measures, which fail to provide fine-grained insights into dataset characteristics and model performance trends~\citep{liang2023holistic}.
One approach that the community has taken to mitigate this problem is to look at instance-level predictions and construct coresets, where each individual dataset in a benchmark is subsampled using various heuristics~\citep{rodriguez-etal-2021-evaluation, perlitz-etal-2024-efficient, Zouhar2025HowTS}.
This resulting subset is used as a proxy for the entire benchmark.
Coreset identification, often done using influence functions~\citep{Koh2017UnderstandingBP, Schioppa2021ScalingUI}, has many drawbacks: integration into an already existing evaluation framework is hard, weak statistical signal hinders generalization, and collecting instance-level predictions across many models may be computationally infeasible~\citep{Chatterjee1986InfluentialOH}. 

\begin{figure*}[t!]
\centering
\includegraphics[width=\linewidth]{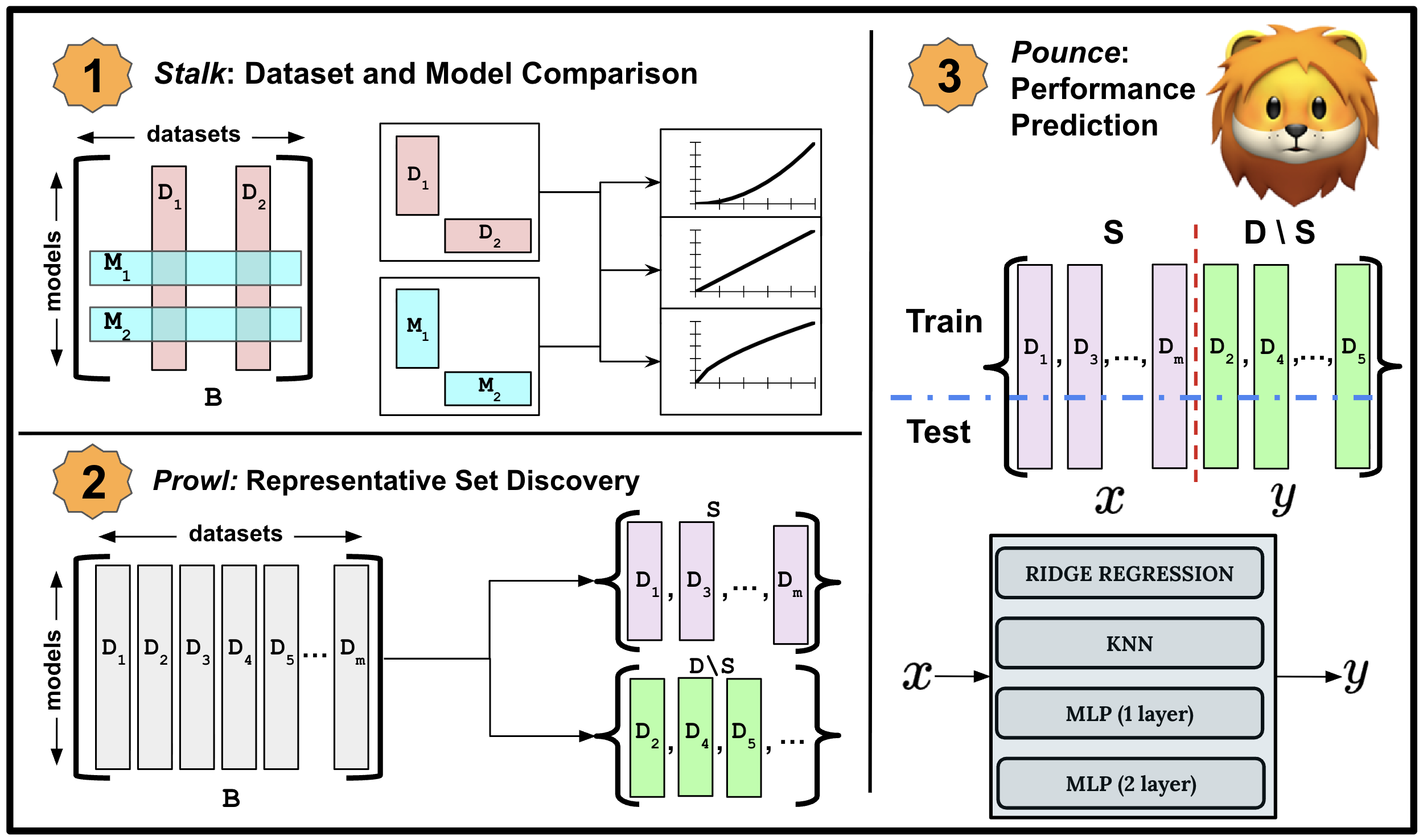}
\caption{An overview of \textbf{SimBA}, our three phase analysis framework. Its three phases are: \texttt{stalk}: dataset \& model comparison, \texttt{prowl}: representative dataset discovery, and \texttt{pounce}: performance prediction.}
\label{fig:framework_overview}
\end{figure*}

Our work looks to uncover a more structured and simplified understanding of benchmarks through an analysis of the datasets and models directly from the performance matrix \emph{without} collecting any instance-level predictions.
Our framework to \textbf{Sim}plify \textbf{B}enchmark \textbf{A}nalysis is called \textbf{SimBA} and has three phases: 
\begin{enumerate}[noitemsep, topsep=0pt]
    \item \texttt{\textbf{Stalk}}: Analyzing relationships between datasets and measuring how models relate to one another across a benchmark.
    \item \texttt{\textbf{Prowl}}: Discovering a representative subset of datasets from a benchmark that maintains model order.
    \item \texttt{\textbf{Pounce}}: Predicting model performances using the representative set based on performance patterns.
\end{enumerate}
Using our three phase approach in \cref{fig:framework_overview}, we analyze HELM~\citep{liang2023holistic}, MMLU~\citep{Hendrycks2020MeasuringMM}, and BigBenchLite~\citep{Srivastava2022BeyondTI} and find that both datasets and models correlate well with one another (\cref{sec: dataset_and_model_comparison}).
Motivated by this, we find that:
\begin{enumerate}[noitemsep, topsep=0pt]
    \item We can identify representative subsets $S_{\text{HELM}}$, $S_{\text{MMLU}}$, and $S_{\text{BBL}}$ with just 6.25\% (1/16), 1.7\% (1/58), and 28.4\% (21/74) of datasets respectively that achieve greater than 95\% coverage (\cref{sec: identifying_the_representative_dataset}).
    \item  Our representative subsets preserve model ranks and can predict performance on a held-out set of models with \emph{near zero} error (\cref{sec: performance_prediction}).
\end{enumerate}
Taken together, our three phase analysis, \textbf{SimBA}, can be used directly by language model practitioners and dataset developers alike to improve efficiency and efficacy.

\section{\texttt{Stalk}: Dataset \& Model Comparison}
\label{sec: dataset_and_model_comparison}
A benchmark is represented as a matrix $B \in \mathbb{R}^{m \times d}$, where $m$ is the number of models and $d$ is the number of datasets. 
$B$ can have missing values.
Different datasets evaluate different metrics, which often are scaled differently (\textit{e.g.} classification vs. generation), so we normalize $B$.
For every dataset $D_i$ with random chance performance $x_{random}$, we modify every observation $x_1, \ldots, x_m$ to be:
\begin{equation}
\label{eq: normalization}
    x_j = max\left(0, \dfrac{x_j - x_{random}}{1 - x_{random}}\right)
\end{equation}
Note these new $x_j$ values correspond to percent above random chance.
We use this normalization for all analysis.~\footnote{\citet{Srivastava2022BeyondTI} use the same normalization for BigBench for some of their analyses.} Moreover, all values are further normalized to be within the interval $[0,1]$, where 0 corresponds to random chance and 1 corresponds to the maximum possible performance.

\subsection{Dataset Comparison}
\label{sec: dataset_comparison}
Datasets are the essence of a benchmark and the first step in our recipe is to compare datasets.
For every pair of datasets $D_i$ and $D_j$, we compare their performance numbers to identify how they relate to each other using one of four relationships: \textsc{linear}, \textsc{exponential}, \textsc{power-law}, or \textsc{none}.
To measure this, we learn a multi-variate linear regressor $f$ with parameters $W_{reg} \in \mathbb{R}^{m \times 1}$ and $b \in \mathbb{R}$ to optimize the objective:
\begin{equation}
    B[:, j] = W_{reg}^{T} \cdot B[:, i] + b
\end{equation}
Here, $M[:, i]$ and $M[:, j]$ are the performance numbers across all models for datasets $D_i$ and $D_j$.
We then obtain the $R^2$ value to quantify the amount of variation explained by the regressor. 
This naturally works to establish the strength of a linear relationship.
To measure an \textsc{exponential} or \textsc{power-law} relationship, we apply a transformation to the measurements of one dataset before learning the linear regressor.~\footnote{This step resembles applying a nonlinear kernel (\eg radial-basis function kernel) to an SVM to learn a nonlinear relationship.}
Since we are predicting one dataset from another, we learn a total of \emph{6} regressors: two per type of relationship (\textsc{linear}, \textsc{exponential}, \textsc{power-law}), one were $D_j$ is predicted from $D_i$ and vice-versa.
We choose the one with the largest $R^2$ value.
If that $R^2 < 0.5$ between two datasets, we classify its relationship as \textsc{none}.~\footnote{$R^2 = 0.5$ indicates that only 50\% of the variation is explained by the regressor, which we deem is too low to confidently claim a specific relationship. This threshold is a tunable hyperparameter.}

\subsection{Model Comparison}
\label{sec: model_comparison}
We compare models across a benchmark in the same way as datasets: for every pair of models $M_1$, $M_2$, we follow the procedure in~\cref{sec: dataset_comparison} to classify whether the relationship is one of four types: \textsc{linear}, \textsc{exponential}, \textsc{power-law}, or \textsc{none}.
Crucially, we only use performance numbers across the benchmark to make this classification.~\footnote{We explicitly do not use any information about the model architecture, size, or family, but one could to improve upon the model comparison phase.}
Identical to dataset comparison, we learn \emph{6} regressors and choose the relationship with the largest $R^2$ value. We classify the relationship between two models to be \textsc{none} if the best regressor results in an $R^2 < 0.5$, just as in~\cref{sec: dataset_comparison}.
\section{\texttt{Prowl}: Representative Dataset Discovery}
\label{sec: identifying_the_representative_dataset}
Equipped with an understanding of a benchmark, our next step is to identify a target. 
More specifically, we want to discover a representative subset of a benchmark $S$ from the performance matrix $B \in \mathbb{R}^{m \times d}$ alone, where $m$ is the number of models and $d$ is the total number of datasets. 

\subsection{Dataset Similarity}
To discover a representative subset, we need a method to determine whether a dataset is redundant.~\footnote{If there exists a set of datasets \{$D_1, \ldots, D_k$\} that are redundant, we need to keep only one in our representative set.}
We compare datasets based on model performance patterns alone using 9 similarity measures: three correlations (\textsc{pearson}, \textsc{spearman}, \textsc{kendall-tau}) and six similarities (\textsc{cosine}, \textsc{manhattan}, \textsc{euclidean}, \textsc{minkowski (p=3)}, \textsc{Wasserstein}, and \textsc{Jensen-Shannon}).~\footnote{All similarities are computed as: \textsc{sim} = 1 - \textsc{distance} or \textsc{sim} = $\exp^{\textsc{distance}}$ based on whether \textsc{distance} is bounded. See~\cref{sec: dataset_similarity_metrics} for more details.}
We compute similarities between all pairs of datasets, resulting in a matrix $C_{\textsc{sim}} \in \mathbb{R}^{d \times d}$, where each entry is a similarity score based on a similarity measure \textsc{sim}. 

\subsection{Discovering a Representative Subset}
First, we define a proxy metric for the coverage of a candidate representative set $S$.
The \textsc{proxy\_coverage} ($\delta$) of $S$ under a similarity measure \textsc{sim} is computed as:

\begin{equation}
\delta(S, \textsc{sim}) = \frac{\sum_{i \in D} \lambda_i}{|\mathcal{D}|}
\end{equation}
where $\lambda_i$ is defined as:

\begin{equation}
\lambda_i = 
\begin{cases}
1, & \text{if } i \in S \\
\max_{j \in S} C_{\textsc{sim}}[i, j], & \text{otherwise}
\end{cases}
\end{equation}

Additionally, we need to define the \textsc{coverage\_gain} ($\Psi$) of a set $S$ when a dataset $D_i$ is added under a similarity measure \textsc{sim}.
This is used as a heuristic to measure how much \textsc{proxy\_coverage} ($\delta$) is gained when adding a new dataset $D_i$ to $S$:
\begin{equation}
   \Psi(S, D_i) = \delta(\{S, D_i\}, \textsc{sim}) - \delta(S, \textsc{sim})
\end{equation}
Equipped with these measures, we propose~\cref{alg:stalk} to discover a representative subset $S$ from a collection of datasets $\{D_1, \ldots D_d\}$. 
While our implementation supports beam search for representative dataset discovery, empirical evaluation showed that larger beam widths (5, 10, 20) provided no meaningful improvement over the greedy approach across HELM, MMLU, and BigBenchLite.~\footnote{$\gamma$ is a \textsc{proxy\_coverage} threshold set by the user. A value of 1 would entail iteratively building $S$ until it contains all the datasets in the benchmark $\{D_1, \ldots D_d\}$.}

\begin{algorithm}[H]
\raggedright
\caption{Representative Dataset Discovery}
\begin{algorithmic}[1]
\label{alg: rep_subset}
\INPUT{$C_{\textsc{sim}} \in \mathbb{R}^{d \times d}$, $\gamma$, $\mathcal{D} = \{D_1, \ldots, D_d\}$}
\OUTPUT{$S$} 
\STATE{$S \gets \emptyset$}
\WHILE{${\delta}(S) < \gamma \land |S| \leq d$}
\STATE{$D^{*} \gets \argmax_{D_i \in D \setminus S} \Psi(S, D_i)$}
\STATE{$S \gets S \cup \{D^*\}$}
\ENDWHILE
\STATE{\bfseries return} $S$
\end{algorithmic}
\label{alg:stalk}
\end{algorithm}

\paragraph{Baselines:}
We also evaluate the following baselines as random and simple baselines in dataset selection have been shown to be strong~\citep{diddee-ippolito-2025-chasing}.
\textsc{Random} is a baseline where $S$ is populated with a random dataset iteratively without replacement.
We average across 1000 random runs in our experiments.
\textsc{Greedy Minimum} is a baseline where $S$ is populated with the dataset with the lowest average performance across all models.
\textsc{Greedy Maximum} is a baseline where $S$ is populated with the dataset with the highest average performance across all models.
For both greedy baselines, $S$ is also populated iteratively. 

\subsection{Representative Subset Evaluation}
Using~\cref{alg: rep_subset}, we discover a representative subset $S$ that has $n$ datasets. 
To measure how well $S$ covers the original benchmark $B$, we first find the mean win rate (\textsc{mwr}) of each model as compared to the other models based entirely on $S$:~
\begin{equation}
    \textsc{mwr}(B') = \dfrac{\sum_{\mu \leq m,d \leq n} \mathbb{I}[B'[\mu,d] > B'[\mu',d]]}{m-1}
\end{equation}

Remember that $B' \in \mathbb{R}^{m \times n}$, where $m$ is the number of models and $n$ is the number of datasets in $S$. $B'$ is matrix formed by collecting all performance numbers for all models for the datasets in $S$, so $\textsc{mwr}(B') \in \mathbb{R}^{m}$.
We then compute the \textsc{pearson-correlation} between $\textsc{mwr}(B')$ and the mean-win-rate obtained on the full benchmark, $\textsc{mwr}(B) \in \mathbb{R}^{m}$, to obtain coverage $\eta$:
\begin{equation}
\label{eq:coverage_via_mwr}
   \eta(B') = \textsc{pearson}(\textsc{mwr}(B), \textsc{mwr}(B')) 
\end{equation}

Identifying $S$ is a greedy process involving a similarity function \textsc{sim}.
Since $S$ iteratively increases from having one to two to many datasets, we require an algorithm to identify how good a similarity function \textsc{sim} is in discovering a strong representative subset at every size. 
Additionally, we also require a method to compare two similarity functions $\textsc{sim}_1$ and $\textsc{sim}_2$.
Taking inspiration from the receiver-operating characteristic curve and taking the area under it (\textsc{auroc};~\citet{marcum1960statistical}), we develop our own signed AUC measure called \textit{subset coverage AUC} (\scauc).~\footnote{\citet{subramani-etal-2025-mice} also develop a signed AUC measure to measure the tool-calling utility of LLMs.}
To compute \scauc, we iteratively build $S$ one dataset at a time until we get to the full benchmark ($S = \{D_1, \ldots, D_d\}$).
Starting with the first dataset chosen ($|S| = 1$), we compute $\eta(B')$ using~\cref{eq:coverage_via_mwr}.~\footnote{Note: $\eta(\emptyset)$ is undefined and $\eta(B) = 1$.}
$B'$ is the matrix formed by taking all the performance numbers for all the datasets in $S$ from the original benchmark matrix $B$.
We then construct a curve using the $d$ coverage values ($\eta$) and compute the signed area under that curve.~\footnote{This is signed area because correlations can be negative.}
To compare two similarity functions, we measure their \scauc values and choose the one with a higher value.
\section{\texttt{Pounce}: Performance Prediction}
\label{sec: performance_prediction}
Equipped with the results of the first two phases of our analysis pipeline, we predict the performances directly using a representative subset.
Since the representative subset $S$ is just a subset of the full benchmark $B$, we evaluate whether different regression based approaches can predict $D \setminus S$ (\ie the subset of $D$ that is not in $S$) from $S$ alone.

One general approach for matrix prediction is Singular Value Decomposition (\textsc{svd}; ~\citet{candes2009exact}).
However, \textsc{svd} works only on a partially observed matrix, where rows (models) and columns (datasets) have at least one observation.
In a realistic setting, we want to use our subset $S$ to predict the other dataset performances $\mathcal{D} \setminus S$ on entirely new models, so \textsc{svd} would not be immediately applicable.
As a result, we focus on three regressor types: \textsc{ridge regression}, \textsc{knn regression}, and \textsc{mlp regression}.

\paragraph{Regularized Linear Regression}
\textsc{Ridge regression} uses a linear function with L2 regularization penalty between a model's performance scores on S and its performance on $D \setminus S$. This regularization helps prevent overfitting when training on small representative subsets~\citep{ hoerl1970ridge, hastie2009elements}.

\paragraph{KNN Regression}

\textsc{knn regression} estimates the performance score \( y \) for a dataset by averaging the performance scores of its \( k \) nearest neighbors in feature space as follows:

\begin{equation}
    y = \frac{1}{k} \sum_{i \in \mathcal{N}_k(X)} y_i
\end{equation}
Here \( \mathcal{N}_k(X) \) denotes the set of the \( k \) closest datasets to \( X \) using a chosen distance metric (\eg \textsc{euclidean})~\citep{altman1992introduction}.
We use $k=5$ neighbors, or the size of the training set if smaller than 5. 
Additionally, \textsc{knn regression} does not impose a functional form, allowing it to capture non-linear relationships. 

\paragraph{MLP Regression}

A Multi-Layer Perceptron (\textsc{mlp}) is a feedforward neural network that models non-linear relationships using multiple layers~\citep{Rosenblatt1958ThePA}.
We experiment with two \textsc{mlp} architectures a single hidden layer with 12 neurons and a two-layer architecture with 12 neurons in each hidden layer.
We include \textsc{mlp regression} because it can capture complex non-linear relationships between dataset features and model performance.

\subsection{Performance Prediction Evaluation}

To evaluate how well we can predict performance, mean squared error (\textsc{mse}) is a natural choice.
For a given representative set $S$, we train a regressor to predict performance on the remaining datasets in the benchmark ($\mathcal{D} \setminus S$).~\footnote{$\mathcal{D}$ is the set of datasets in the benchmark $\mathcal{D} = \{D_1, ..., D_d\}$.}
We compute the \textsc{mse} of the regressor on the held-out test set on ($\mathcal{D} \setminus S$).

In our experiments, we build $S$ sequentially, by greedily adding one dataset at a time according to~\cref{alg: rep_subset}.
As a result, we can measure \textsc{mse} at each point for $|S| = 1, ..., |S| = |\mathcal{D}|-1$.~\footnote{When $|S| > |\mathcal{D}|-1$, ($|\mathcal{D} \setminus S|$ = 0).}
This traces an \textsc{mse} curve.
Using a similar approach to tracing the area under the coverage curve like in~\cref{sec: identifying_the_representative_dataset}, we can compute the area under the \textsc{mse} curve, which we term \aucmse.~\footnote{\aucmse is not signed because the minimum error one can get is 0.0, so every point on this curve is in the top right quadrant of a cartesian coordinate plane ($x, y > 0$).}
Note that high values of \aucmse indicate high error because this curve is an error curve \emph{not} a performance curve like other \textsc{auc} curves.
To measure which regressor is best, we compare \aucmse values and choose the method with the lowest \aucmse value.
\section{Experiments}

\begin{figure*}[t]
   \includegraphics[width=\textwidth]{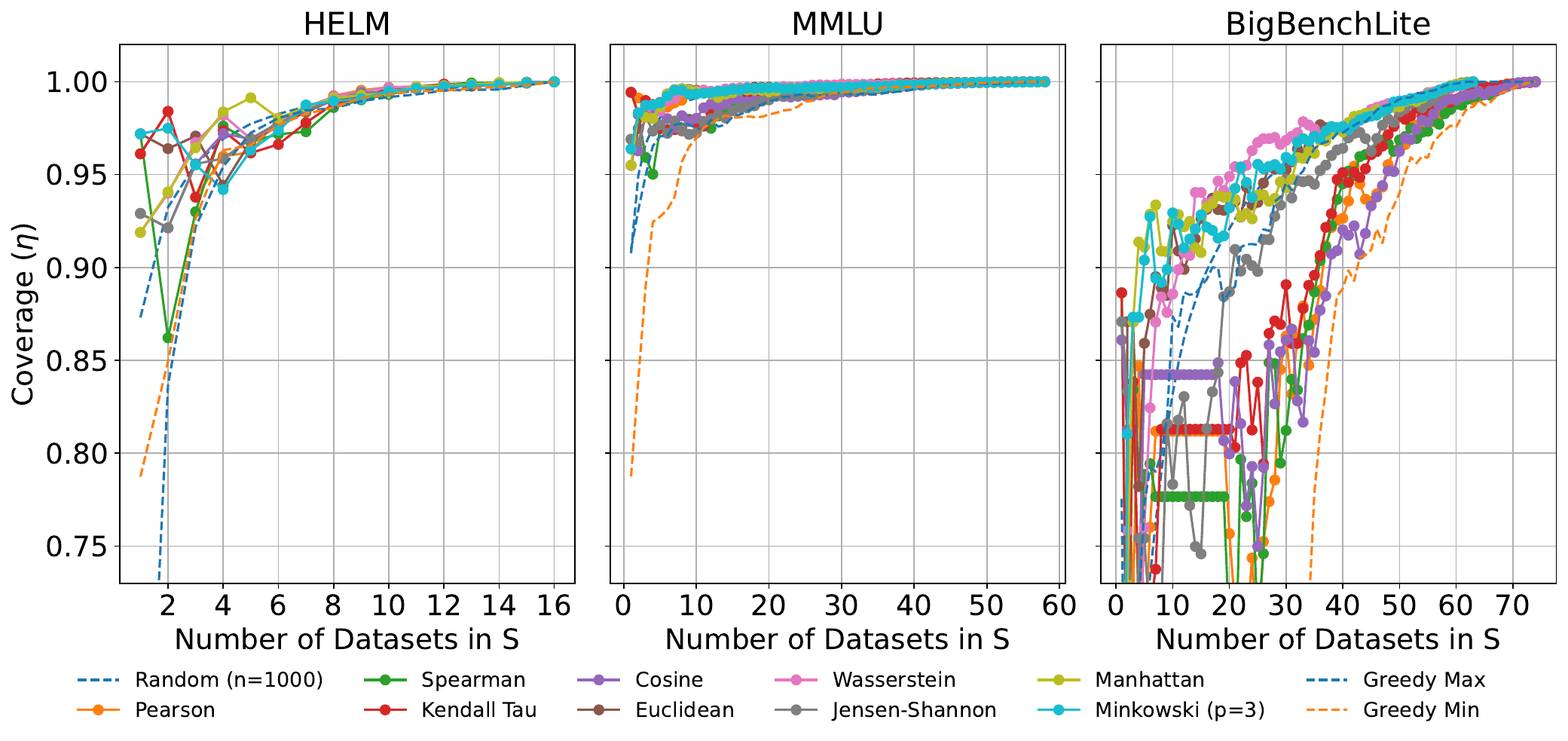}
  \caption{Here, we measure the coverage ($\eta$) across HELM, MMLU, and BigBenchLite as our representative subset $S$ grows. We report performance for all three baselines and all nine similarity measures discussed in~\cref{sec: identifying_the_representative_dataset}.}
  \label{fig:coverage_all_benchmarks}
\end{figure*}
\paragraph{Benchmarks}
For our analysis, we look at three benchmarks: HELM~\citep{liang2023holistic}, MMLU~\citep{Hendrycks2020MeasuringMM}, and BigBenchLite~\citep{Srivastava2022BeyondTI}.
We look at the core scenarios of HELM (17 datasets, 29 models), MMLU (58 datasets, 79 models), and BigBenchLite (74 datasets, 45 models), splitting each benchmark into training and test sets to validate our analysis. 
All datasets are included in both splits, but models are separated across training and test sets randomly with 80\% of models in training and 20\% of the models in test. 
This means that HELM has 23 models in train and 6 models in test, MMLU has 63 models in train and 16 in test, and BigBenchLite has 36 models in train and 9 in test.
For each benchmark, we go through the 3 analysis phases from~\cref{fig:framework_overview}, discuss those results in~\cref{sec: results}, and expand the analysis to look at robustness in~\cref{sec: analysis}.

\section{Results}
\label{sec: results}

\subsection{\texttt{Stalk}: Dataset \& Model Comparison}
\paragraph{Dataset Comparison:} We measure how datasets relate to one another using the methodology in~\cref{sec: dataset_comparison}.~\cref{tab:dataset_comparison} shows that for HELM, only 5\% of the pairs of datasets have a \textsc{linear} relationship, while 37.5\% lack any relationship, indicating minor redundancy among datasets.

For MMLU, 4.1\% of dataset pairs have a \textsc{linear} relationship and only 7.7\% lack a relationship, suggesting that most datasets are predictive of one another through non-linear relationships (\textsc{exponential}: 39.3\%, \textsc{power-law}: 49.0\%).
For BigBenchLite (BBLite), 27.9\% of dataset pairs have a \textsc{linear} relationship while 61.6\% show no relationship at all, indicating that BBLite has the most diverse and independent datasets among the three benchmarks. 
Taken together, we suspect that finding a small representative dataset would be most difficult for BBLite.

\begin{table}[t]
\centering
\begin{tabular}{@{}lccc@{}}
\toprule
Relationships & HELM & MMLU & BBLite\\
\midrule
\textsc{linear}      & 6   & 67  & 754\\
\textsc{exponential} & 25  & 649 & 139\\
\textsc{power-law}   & 44  & 810 & 143\\
\textsc{none}        & 45  & 127 & 1665\\
\midrule
Total       & 120 & 1653& 2701\\
\bottomrule
\end{tabular}
\caption{Dataset comparison results with counts between each pair of datasets and their classifications as \textsc{linear}, \textsc{exponential}, \textsc{power-law}, and \textsc{none} based on the highest $R^2$ value. See~\cref{sec: dataset_comparison} for details.}
\label{tab:dataset_comparison}
\end{table}

\begin{table}[t]
\centering
\begin{tabular}{@{}lccc@{}}
\toprule
Relationships & HELM & MMLU & BBLite \\
\midrule
\textsc{linear}      & 20  & 290  & 75\\
\textsc{exponential} & 72  & 1258 & 41\\
\textsc{power-law}   & 294 & 1302 & 549\\
\textsc{none}        & 20  & 231  & 325\\
\midrule
Total       & 406 & 3081 & 990\\
\bottomrule
\end{tabular}
\caption{Model comparison results with counts between each pair of datasets and their classifications as \textsc{linear}, \textsc{exponential}, \textsc{power-law}, and \textsc{none} based on the highest $R^2$ value. See~\cref{sec: model_comparison} for details.}
\label{tab:model_comparison}
\end{table}

\paragraph{Model Comparison:} We measure how models relate to one another via the method in~\cref{sec: model_comparison}.
\cref{tab:model_comparison} shows that for HELM, only 4.9\% of model pairs have \textsc{linear} relationships, with the majority showing \textsc{power-law} relationships (72.4\%). MMLU demonstrates more complex model relationships with 9.4\% \textsc{linear}, 40.8\% \textsc{exponential}, and 42.3\% \textsc{power-law} relationships.
BigBenchLite shows 7.6\% \textsc{linear} relationships and 55.5\% \textsc{power-law} relationships. Notably, all three benchmarks have relatively few model pairs with no discernible relationship (HELM: 4.9\%, MMLU: 7.5\%, BigBenchLite: 32.8\%).
BigBenchLite again has the highest rate of no relationships further hinting that discovering a representative subset for BigBenchLite may be the most difficult.

\begin{table*}[h!]
\centering
\begin{tabular}{@{}lcccccc@{}}
\toprule
                            & \multicolumn{2}{c}{HELM} & \multicolumn{2}{c}{MMLU} & \multicolumn{2}{c}{BigBenchLite} \\
Similarity Methods                     & \scauc $(\uparrow)$ & $|S^*| (\downarrow)$  & \scauc $(\uparrow)$ & $|S^*| (\downarrow)$  & \scauc $(\uparrow)$ & $|S^*| (\downarrow)$  \\
\midrule
\textsc{random} (n = 1000) & 0.980 & 2.5  & 0.994 & 2.3  & 0.928 & 25.2 \\
\textsc{greedy minimum} & 0.967& 4 & 0.980& 8 & 0.607& 51 \\
\textsc{greedy maximum} & 0.957& 4 & 0.988& 4 & 0.933& 33 \\
\midrule
\textsc{pearson}         & 0.984 & \textbf{1} & \textbf{0.997}& \textbf{1} & 0.886& 42 \\
\textsc{spearman}        & 0.975 & \textbf{1} & 0.992& \textbf{1} & 0.884& 41 \\
\textsc{kendall-Tau}     & 0.983 & \textbf{1} & 0.994& \textbf{1} & 0.899& 40 \\
\textsc{cosine}          & 0.981 & 3 & 0.992& \textbf{1} & 0.896& 48 \\
\textsc{manhattan}       & \textbf{0.985} & 3 & 0.996& \textbf{1} & 0.945& 32 \\
\textsc{euclidean}       & 0.984 & \textbf{1} & 0.996& \textbf{1} & 0.943& 27 \\
\textsc{minkowski (p=3)} & 0.983 & \textbf{1} & 0.996& \textbf{1} & \textbf{0.950}& 22 \\
\textsc{wasserstein}     & 0.983 & 3 & \textbf{0.997}& \textbf{1} & 0.943& \textbf{21} \\
\textsc{jensen-shannon}  & 0.980 & 3 & 0.991& \textbf{1} & 0.903& 36 \\
\bottomrule
\end{tabular}
\caption{Performance of our three baselines and seven similarity measures on the identification of a representative subset task for HELM, MMLU, and BigBenchLite. \scauc is the area under the coverage curves present in~\cref{fig:coverage_all_benchmarks}. $S^*$ is the smallest subset that achieves $\eta(S^*) = 0.95$. \textbf{Bold} indicates the best performing system for each metric.}
\label{tab:scauc_results}
\end{table*}
\subsection{\texttt{Prowl}: Coverage Analysis}
\label{sec: prowl_coverage_analysis}
Our goal is to determine the minimum number of datasets needed to achieve a specific coverage level $\eta$ on a specific benchmark.
Since we experiment with nine similarity functions, we want to identify the best similarity measure for this task.
To do this, we first compare every pair of datasets $D_i$, $D_j$ using each of the similarity functions defined in~\cref{sec: identifying_the_representative_dataset}. This results in a similarity matrix $C_{\textsc{sim}} \in \mathbb{R}^{d \times d}$ for each similarity function.~\footnote{$C_{\textsc{sim}}$ could be an upper (or lower) triangular matrix because our similarity functions are symmetric.}
On each similarity matrix $C_{\textsc{sim}}$, we apply our coverage algorithm (\cref{alg: rep_subset}) using its respective similarity function and construct $S$.
We measure coverage ($\eta$) using~\cref{eq:coverage_via_mwr} at each iteration as $S$ is being constructed until $S = \mathcal{D}$.
This traces a coverage curve and we measure the area under this coverage curve and report this \scauc value in~\cref{tab:scauc_results}. 
We also report the size of the \emph{smallest} representative subset $S*$ that achieves a coverage $\eta \geq 0.95$.

We find that we can achieve coverage levels of at least 95\% with just 6.25\% (1/16), 1.7\% (1/58), and 28.4\% (21/74) of the datasets for HELM, MMLU, and BigBenchLite respectively. This represents a substantial efficiency gain: particularly for MMLU and HELM, where a single well-chosen dataset can effectively represent nearly the entire benchmark for model ranking purposes.
Additionally,~\cref{tab:scauc_results} shows that the choice of similarity measure has varying effects across benchmarks. For HELM and MMLU, most similarity measures perform similarly well, with several achieving the optimal single-dataset representative subset. However, for BigBenchLite, there is more variation in performance, with \textsc{wasserstein} achieving the best results (|S*| = 21) and several measures like \textsc{pearson} and \textsc{spearman} requiring substantially more datasets (|S*| = 42 and 41 respectively).

Finally, using~\cref{alg: rep_subset} with most similarity measures outperforms all baselines on average across all three benchmarks, though the improvement is less pronounced for BigBenchLite due to its more diverse dataset composition.
We also compute how often using~\cref{alg: rep_subset} with a similarity measure outperforms the \textsc{random} baseline.
See~\cref{tab:performance_comparison_vs_random} for details on the proportion of times a system outperforms the random baseline across 1000 random runs.

\begin{table}[h!]
\centering
\begin{tabular}{@{}lccc@{}}
\toprule
           & \multicolumn{3}{c}{\aucmse ($\downarrow$)} \\
Regressor & HELM & MMLU & BBLite \\
\midrule
\textsc{ridge}            & 0.005& \textbf{0.002}& \textbf{0.002}\\
\textsc{knn} & \textbf{0.004}& \textbf{0.002}& \textbf{0.002}\\
\textsc{mlp} (1 layer)    & 0.081& 0.110& 0.006\\
\textsc{mlp} (2 layer)    & 0.031& 0.081& 0.006\\
\bottomrule
\end{tabular}
\caption{Performance as measured by \aucmse across HELM, MMLU, and BigBenchLite. We use \textsc{Minkowski (p=3)} as the similarity measure \textsc{sim} to identify representative subsets $S$ for our four regressors: \textsc{ridge}, \textsc{knn}, \textsc{mlp} (1 layer), and \textsc{mlp} (2 layer). \aucmse is the area under the curves in the top row of~\cref{fig:prediction_all_benchmarks_noise}. \textbf{Bold} indicates the best performing method for each benchmark.}
\label{tab:perf_results}
\end{table}

\subsection{\texttt{Pounce}: Performance Prediction}

Our goal in this phase is to validate that representative datasets enable accurate performance prediction. Having identified efficient representative subsets in phase II (\texttt{prowl}), we now assess whether performance on these subsets can predict performance on the remaining datasets.
We first split our models into training (80\%) and test (20\%) sets. Using only the models in the training set, we identify representative datasets at 80\% coverage and train four performance predictors: \textsc{ridge regression}, \textsc{knn regression}, and \textsc{mlp regression} with one and two layers. 
We then evaluate these predictors on the test set of models, measuring their ability to predict scores (\textsc{mse}) for the remaining datasets based solely on performance patterns observed in the representative subset.

\begin{figure*}[t!]
  \includegraphics[width=\textwidth]{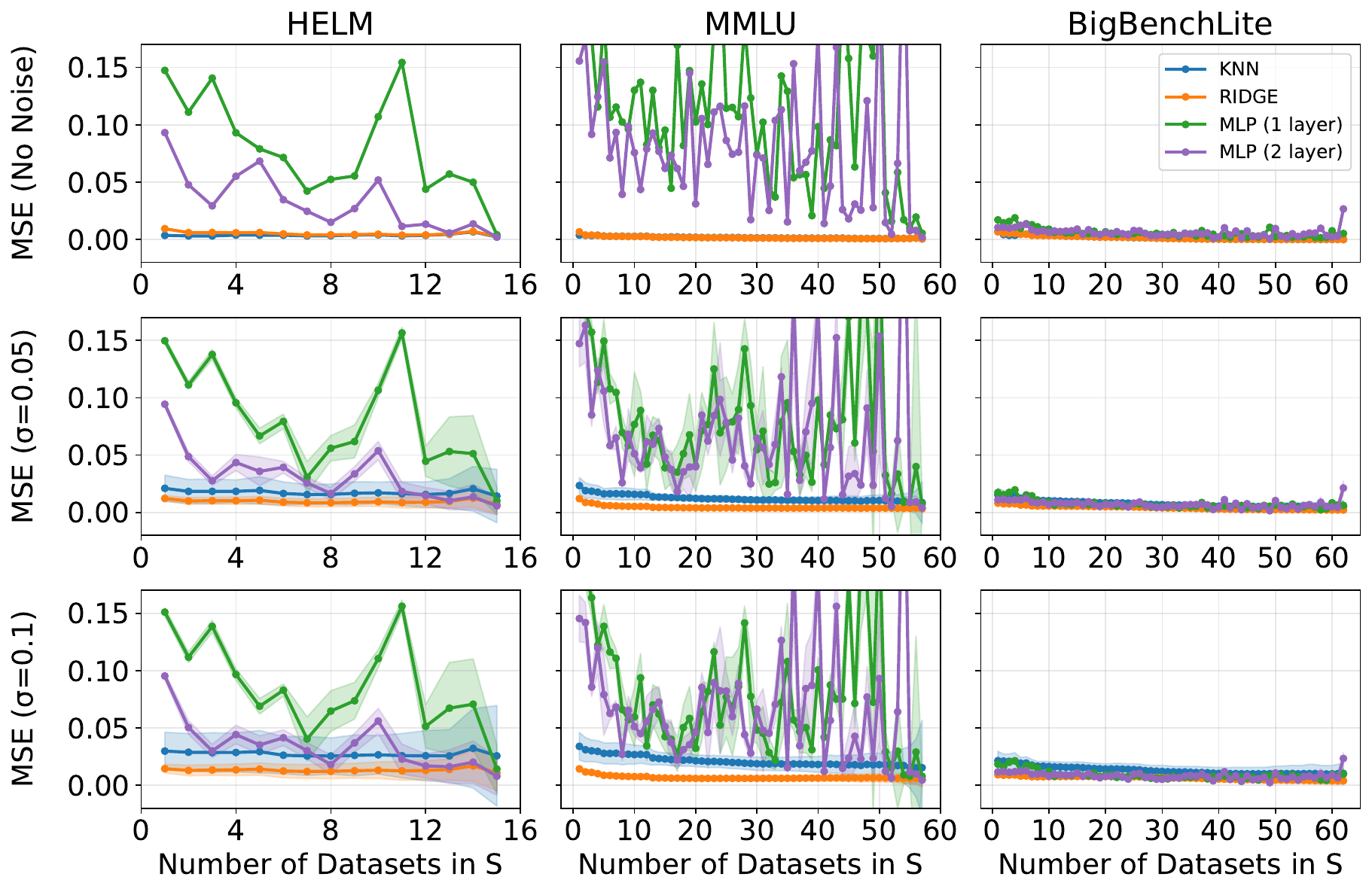}
  \caption{We measure the mean squared error (MSE) on a held-out test set of models of regressors trained on a representative subset $S$ using \textsc{Minkowski (p=3)} as \textsc{sim}. This is measured for HELM, MMLU, and BigBenchLite for the three regressors we experiment with in~\cref{sec: performance_prediction}. Additionally, we repeat this experiment after adding two magnitudes of noise to the training set of $B$ according to~\cref{sec: robustness} ($\sigma = 0.05$ and $\sigma = 0.1$). Lower scores are better.} 
  \label{fig:prediction_all_benchmarks_noise}
\end{figure*}

As shown in~\cref{tab:perf_results},  both \textsc{ridge regression} and \textsc{knn regression} perform exceptionally well across all three benchmarks, achieving \emph{near zero} \aucmse values. \textsc{ridge regression} achieves (0.005, 0.002, 0.002) and \textsc{knn regression} achieves (0.004, 0.002, 0.002) for HELM, MMLU, and BigBenchLite respectively.  Meanwhile,~\cref{fig:prediction_all_benchmarks_noise} shows that prediction error generally decreases as the representative subset size increases, but with diminishing returns. The MLP models consistently underperform, with the single-layer MLP showing particularly poor results on HELM and MMLU.
Additionally, in~\cref{fig:prediction_all_benchmarks_noise}, we find that \textsc{knn regression} achieves negligible error with just \emph{one} dataset on both HELM and MMLU.
This could suggest that these benchmarks are saturated.

For MMLU, all four regressors maintain consistently low error rates across different subset sizes, with \textsc{ridge} and \textsc{knn} maintaining the lowest error rates throughout. 
This complements the analysis presented in~\cref{sec: dataset_comparison}, where MMLU showed the highest proportion of inter-dataset relationships, making it the most predictable benchmark.

\section{Analysis}
\label{sec: analysis}

\subsection{Is \texttt{Pounce} robust to noise?}
\label{sec: robustness}
Our evaluation framework relies on point estimates of the performances of models on individual datasets.
As such, robustness to noise is a critical consideration when evaluating different approaches of performance prediction.
We evaluate the robustness of performance prediction by perturbing the training set of the benchmark matrix $B_{train}$ as $B_{train}' = B_{train} +\mathcal{N}(\mu,\sigma^2)$ where $\mu$ is mean and $\sigma$ is the noise level parameter. For all noise perturbations we use $\mu=0$ and $\sigma^2 = 0.05$ or $\sigma^2 = 0.1$. 
This evaluation framework enables us to quantify the stability of our methods under varying noise conditions.

As shown in~\cref{fig:prediction_all_benchmarks_noise}, generally, error rates for all methods increase with greater noise. However, we observe that both \textsc{ridge} and \textsc{knn regression} achieve low error rates under both noise conditions, across all three benchmarks (\aucmse < 0.03). \textsc{ridge} consistently maintains \aucmse values of 0.013 for all benchmarks and noise conditions.
Predictably, the \textsc{mlp} systems are the least robust, but maintain similar error rates to the no noise version.
In other words, the \textsc{mlp} systems just are not great at performance prediction.

\newcommand{\rot}[1]{\rotatebox[origin=c]{90}{#1}}
\begin{table}[]
\centering
\begin{tabular}{@{}clccc@{}}
\toprule
           & & \multicolumn{3}{c}{\aucmse ($\downarrow$)} \\
& Regressor & HELM & MMLU & BBLite \\
\midrule
\multirow{4}{*}{\rot{$\sigma = 0.05$}} & \textsc{ridge} & $\textbf{0.010}$ & $\textbf{0.004}$ & $\textbf{0.004}$\\
& \textsc{knn} & $0.017$ & $0.012$ & $0.013$\\
& \textsc{mlp} (1 layer) & $0.081$ & $0.079$ & $0.007$\\
& \textsc{mlp} (2 layer) & $0.031$ & $0.068$ & $0.007$\\
\midrule
\multirow{4}{*}{\rot{$\sigma = 0.1$}} & \textsc{ridge} & $\textbf{0.013}$ & $\textbf{0.007}$ & $\textbf{0.006}$\\
& \textsc{knn} & $0.027$ & $0.021$ & $0.013$\\
& \textsc{mlp} (1 layer) & $0.087$ & $0.078$ & $0.009$\\
& \textsc{mlp} (2 layer) & $0.033$ & $0.067$ & $0.008$\\
\bottomrule
\end{tabular}
\caption{Performance as measured by \aucmse across HELM, MMLU, and BigBenchLite in the presence of noise (\cref{sec: robustness}). We use \textsc{Minkowski (p=3)} as the similarity measure \textsc{sim} to identify representative subsets $S$ for our four regressors: \textsc{ridge}, \textsc{knn}, \textsc{mlp} (1 layer), and \textsc{mlp} (2 layer). \aucmse is the area under the curves in~\cref{fig:prediction_all_benchmarks_noise}. \textbf{Bold} indicates the best performing method for each benchmark.}
\label{tab:perf_noise_results}
\end{table}

\subsection{Is \texttt{Pounce} robust to varying data splits?}
\label{sec: robust_data_splits}
Here, we look at how different subsets or selections of data change the \texttt{pounce} stage.
This is related to the noise level analysis in~\cref{sec: robustness}, but requires its own investigation.
The former approximates uncertainty for individual observations, whereas here, we care about measuring the variance of \texttt{pounce} when the training and test splits vary.

To keep the test set uncontaminated, we look at just the training set.
On this, we perform k-fold cross validation and measure performance via \aucmse.~\footnote{We use 5-fold cross validation for HELM and 10-fold for MMLU and BigBenchLite.}.
We observe low standard deviations across different training splits (< 0.01) indicating that performance prediction is stable.
Intuitively, the simpler models, \textsc{ridge} and \textsc{knn}, are the most stable as they are the least prone to overfitting.
Taken together, our analysis suggests that similar models tend to perform similarly on related tasks, hinting that local neighborhood relationships are effective for \texttt{pounce}.
Overall our error rates remain small and are robust to perturbations or changes in both the training and evaluation sets.
\section{Related Work}
The NLP community increasingly evaluates on large benchmarks~\citep{Srivastava2022BeyondTI, alpaca_eval, liang2023holistic}.
Some approaches attempt to make evaluation more efficient by doing instance-level reduction~\citep{vivek-etal-2024-anchor, Polo2024tinyBenchmarksEL, perlitz-etal-2024-efficient}.
\citet{MagnussonDataDecide2025} look at a loosely related problem and use small scale experiments for pretraining data selection.
They, too, require instance-level information and are focused on analyzing how training on different subsets of pretraining data affect performance, rather than discovering a representative set of pretraining corpora.
We differ from all of the above.
Our work focuses on aggregate metrics, specifically not requiring access to any instance-level information.

As mentioned earlier, most work looks at identifying coresets of benchmarks at the instance-level and there exist numerous methods to do this across many different application areas~\citep{Lewis1994HeterogeneousUS, Killamsetty2021GRADMATCHGM, Paul2021DeepLO, Moser2025ACS}.
\citet{ye-etal-2023-predictable} is one of the few works that, like us, goes beyond instance-level work and tackles the performance prediction task using simple regressors on BigBench. 
However, their approach is not as lightweight as ours because they requires features of the model and datasets for accurate prediction.

From the field of psychometrics and measurement theory, there exists the idea of convergent and divergent validity~\citep{campbell1959convergent}, which helps contextualize our proposed framework. 
Convergent validity suggests that metrics measuring similar underlying constructs should correlate highly with each other. Conversely, discriminant validity indicates that metrics capturing fundamentally different aspects should show minimal correlation. \citet{xiao-etal-2023-evaluating-evaluation} propose MetricEval, a framework motivated by measurement theory to conceptualize and evaluate the reliable and valid of natural language generation metrics. 

\section{Conclusion}
We propose a three phase approach to \textbf{Sim}plify \textbf{B}enchmark \textbf{A}nalysis called \textbf{SimBA}: \texttt{stalk} (dataset \& model comparison), \texttt{prowl} (representative set discovery), and \texttt{pounce} (performance prediction).
Using our approach, we analyze the HELM, MMLU, and BigBenchLite benchmarks.
Our analysis shows that models and datasets alike correlate well with one another (\texttt{stalk}).
Additionally, using~\cref{alg: rep_subset}, we can identify representative subsets with 1, 1, and 21 datasets respectively that achieve a greater than 95\% coverage (\texttt{prowl}).
These representative sets preserve the original model ranks on the benchmark and can be used to predict performance on held-out models with \emph{negligible} error (\texttt{pounce}).
Furthermore, \textbf{SimBA} can be used by LM practitioners and dataset developers directly to reduce evaluation costs and validate dataset uniqueness.
\section{Limitations}
\paragraph{Dataset \& Model Comparison}
Our analysis assumes that the relationships between datasets are sufficiently stable over time. As models continue to improve and scale, the nature of these relationships may evolve, potentially requiring periodic reassessment of representative subsets. The approach provides a snapshot analysis based on current model performance matrices, but doesn't account for how these relationships might change with fundamentally new architectures or training paradigms.

Moreover, the presented analyses requires having a sufficient number of models evaluated on the benchmarks. 
If the available models lack in diversity in their underlying architectures, training data, or training methodologies, the identified relationships may not generalize to future models. 

\paragraph{Identifying the Representative Dataset}
Our method is a greedy approach that iteratively chooses datasets such that \textsc{proxy\_coverage} increases. A different, albeit more expensive, approach could exhaustively identify the best combination of datasets using a better search algorithm.
We experimented with adapting to beam search, but found no significant improvement, perhaps because \textsc{proxy\_coverage} is correlated, but can be slightly disconnected with \textsc{coverage} depending on the similarity function used.

\paragraph{Performance Prediction}
Although \textsc{knn} performs reasonably well, its \aucmse values are consistently 2-3 times higher than \textsc{ridge} in the presence of noise.
In these settings the \textsc{mlp} models perform the worst with \aucmse values 5-10 times higher than \textsc{ridge}, possibly due to overfitting on the limited training data.
Models trained with better regularization on more data would have greater stability and be less prone to overfitting, so this is something we recommend for practitioners using \textbf{SimBA}.

\paragraph{Overall Risks}
Our evaluation framework offers a three stage approach to better understand a benchmark.
Although representative sets get high coverage, there could be cases when the representative set gets high coverage by chance.
In this case, it would be risky to make major decisions that affect users based on a small sample of data.
\section{Ethical Considerations}

Since SimBA does not involve training generative models, the primary ethical concerns center on potential misuse of our framework's insights and the risk of overconfidence in representative subset evaluations.

Our finding that small representative subsets can achieve high coverage creates opportunities for manipulation. Model developers could strategically evaluate only on datasets where their models perform well, then use \textbf{SimBA} to claim coverage over the entire benchmark without actually testing on challenging datasets. This could mislead the research community and downstream users about true model capabilities. Similarly, the choice of similarity measure presents another avenue for selective reporting, as our analysis shows that different similarity functions (\textsc{pearson}, \textsc{minkowski (p=3)}, \textsc{wasserstein}, etc.) can yield different representative subsets and coverage results.

To aid in mitigating these concerns, we recommend transparent reporting of representative set selection methodologies, evaluation across multiple correlation methods, and validation on diverse datasets.
SimBA does not aim to replace comprehensive evaluation, especially for high-stakes deployments. Rather, it serves as a supplementary tool for understanding benchmark structure and improving evaluation efficiency in appropriate contexts.
\section*{Acknowledgments}
We thank Iz Beltagy, Pradeep Dasigi, Dirk Groeneveld, and Alexis Ross for feedback on very early scoping of this work. Additionally, we thank Harshita Diddee, Athiya Deviyani, and members of MD's R3Lit lab for helpful discussions and feedback on later versions of the work. This research was in part supported by the National Institute of Standards and Technology (\url{ror.org/05xpvk416}) under Federal Award ID Number 60NANB24D231 and Carnegie Mellon University (\url{https://ror.org/05x2bcf33}) AI Measurement Science and Engineering Center (AIMSEC). AG is supported by the NSF CSGrad4US Fellowship.

\bibliography{fullbib}

\onecolumn
\clearpage
\newpage
\appendix
\label{sec:appendix}
\section{Dataset Similarity Metrics}
\label{sec: dataset_similarity_metrics}
Here are more details about the dataset similarity metrics used in our analysis.
Consider a pair of datasets $(D_i, D_j)$; we use seven similarity measures \textsc{sim}.

\paragraph{\textsc{pearson correlation}:} Measures linear relationships between dataset performance vectors but is sensitive to outliers and assumes linearity. We compute the \textsc{pearson correlation}:
\begin{equation}
\rho_{D_i, D_j} = \frac{\sum (R_{m,D_i} - \bar{R}_{D_i})(R_{m,D_j} - \bar{R}_{D_j})}
    {\sqrt{\sum (R_{m,D_i} - \bar{R}_{D_i})^2 \sum (R_{m,D_j} - \bar{R}_{D_j})^2}}
\end{equation}
where $R_{m,D_i}$ is the performance of model $m$ on dataset $D_i$ and $\bar{R}_{D_i}$ is the average performance across all models for dataset $D_i$,

\paragraph{\textsc{spearman correlation}:} A ranked correlation that handles non-linear monotonic relationships but is sensitive to small perturbations that flip ranks.

\begin{equation}
    \rho_s = 1 - \frac{6 \sum d_i^2}{n(n^2 - 1)}
\end{equation}
where \( d_i \) is the rank difference between model performances on datasets \( D_i \) and \( D_j \),

\paragraph{\textsc{kendall-tau correlation \cite{kendall1938new}}:} Another ranked correlation that measures agreement in the orderings of data but also sensitive to small perturbations.

\begin{equation}
    \tau = \frac{C - D}{C + D}
\end{equation}
where \( C \) represents concordant model rankings across two datasets, and \( D \) represents discordant rankings.
\paragraph{\textsc{cosine similarity}:} Measures the cosine of the angle between performance vectors, capturing directional similarity regardless of magnitude differences between datasets.
\begin{equation}
   \textsc{cosine\_similarity}(D_i, D_j) = \frac{B[:, i] B[:, j]}{||B[:, i]|| ||B[:, j]||}
\end{equation}

\paragraph{\textsc{lp norm \cite{rudin1987real} based similarities}:}
We define a family of similarity measures based on Lp norms with exponential normalization:
\begin{align}
\textsc{lp\_similarity}(D_i, D_j) &= \exp\left(-||B[:, i] - B[:, j]||_p\right)
\end{align}
We employed various distance-based similarity measures using exponential normalization to capture different aspects of performance similarity between datasets. We specifically use L1 (\textsc{manhattan}), L2 (\textsc{euclidean}), and L3 (\textsc{minkowski}) norms. The exponential normalization ensures all similarities are bounded in (0,1], with identical performance patterns yielding similarity 1 and increasingly dissimilar patterns approaching 0.

\paragraph{\textsc{wasserstein similarity \cite{mallows1972note}}:} Measures the minimum "cost" of transforming one performance distribution into another, capturing both shape and statistical differences. We also exponentially normalize this such that similarities are bounded in (0, 1].
\begin{equation}
\textsc{wasserstein\_similarity}(D_i, D_j) = \exp\left(\frac{-W_1(B[:, i], B[:, j])}{\max_{k,l} W_1(B[:, k], B[:, l])}\right)
\end{equation}

\paragraph{\textsc{jensen-shannon similarity}:} 
\begin{equation}
\textsc{jensen\_shannon\_similarity}(D_i, D_j)= 1 - \sqrt{\frac{D_{KL}(P_i||M) + D_{KL}(P_j||M)}{2}}
\end{equation}
Here \(M = \frac{1}{2}(P_i + P_j)\) and \(P_i, P_j\) are normalized distributions of \(B[:, i], B[:, j]\). Jensen-Shannon similarity provides a symmetric measure based on information theory that quantifies dataset distributional differences.

\section{Proportions Better than Random}
To measure how well each system (other than random) fares across the 1000 random runs of the \textsc{random} baseline, we measure the proportion of the 1000 random runs that each system does as well or better than.
We measure this across two metrics on all three benchmarks.
The first metric, "AUC", is measured by \scauc. 
The second metric, "Max2", looks at the \scauc up until a representative dataset $S$ is discovered that achieves at least 95\% coverage (S*). 
This is compared to the \textsc{random} baseline for the same number of datasets (|S*| under a similarity function \textsc{sim}) across all 1000 runs.
Note that \scauc cannot be computed if |S*|=1, so if |S*|=1, we consider the first two datasets here.
These results are below in~\cref{tab:performance_comparison_vs_random} as proportions.
We find that \textsc{greedy minimum} and \textsc{greedy maximum} both perform worse than \textsc{random}.
On MMLU, all similarity measures outperform \textsc{random}, but on HELM and BigBenchLite, only about half the systems outperform \textsc{random} on \scauc on average.
Our representative dataset discovery algorithm generally outperforms \textsc{random} early on until S* is discovered regardless of similarity function, with most systems strongly outperforming \textsc{random} on "Max2."

\begin{table*}[h]
\centering
\begin{tabular}{@{}lcccccc@{}}
\toprule
                            & \multicolumn{2}{c}{HELM} & \multicolumn{2}{c}{MMLU} & \multicolumn{2}{c}{BigBenchLite} \\
Similarity Methods          & AUC $(\uparrow)$ & Max2 $(\uparrow)$ & AUC $(\uparrow)$ & Max2 $(\uparrow)$ & AUC $(\uparrow)$ & Max2 $(\uparrow)$ \\
\midrule
\textsc{random} (baseline)  & -- & -- & -- & -- & -- & -- \\
\textsc{greedy minimum}     & 0.095 & 0.095 & 0.000 & 0.000 & 0.000 & 0.000 \\
\textsc{greedy maximum}     & 0.030 & 0.028 & 0.002 & 0.223 & 0.521 & 0.504 \\
\midrule
\textsc{pearson}            & 0.626 & {0.970} & \textbf{{0.972}} & \textbf{{1.000}} & 0.068 & 0.086 \\
\textsc{spearman}           & 0.213 & 0.415 & 0.108 & 0.994 & 0.057 & 0.072 \\
\textsc{kendall-tau}        & 0.552 & {0.970} & 0.330 & 0.994 & 0.141 & 0.154 \\
\textsc{cosine}             & 0.427 & 0.409 & 0.124 & 0.880 & 0.115 & 0.132 \\
\textsc{manhattan (L1)}     & \textbf{{0.685}} & 0.501 & 0.825 & 0.924 & 0.727 & 0.849 \\
\textsc{euclidean}          & 0.595 & 0.944 & 0.931 & 0.868 & 0.692 & 0.832 \\
\textsc{minkowski (L3)}     & 0.538 & \textbf{0.973} & 0.939 & 0.965 & \textbf{{0.815}} & \textbf{{0.928}} \\
\textsc{wasserstein}        & 0.581 & 0.507 & {0.962} & 0.965 & 0.692 & 0.791 \\
\textsc{jensen-shannon}     & 0.377 & 0.409 & 0.060 & 0.919 & 0.155 & 0.303 \\
\bottomrule
\end{tabular}
\caption{Proportion of methods that perform better than or equal to \textsc{random} across three evaluation metrics. Values closer to 1.0 indicate better performance relative to \textsc{random}. \textbf{Bold} indicates the best performing method for each metric within each dataset.}
\label{tab:performance_comparison_vs_random}
\end{table*}

\end{document}